\documentclass[10pt,conference,a4paper]{IEEEtran}
\IEEEoverridecommandlockouts
\usepackage{cite}
\usepackage{amsmath,amssymb,amsfonts}
\usepackage{algorithmic}
\usepackage{graphicx}
\usepackage{textcomp}
\usepackage{xcolor}
\usepackage{hyperref}
\usepackage{balance}

\begin{document}

\title{Investigating and Exploiting Image Resolution for Transfer Learning-based Skin Lesion Classification
\thanks{© 2020 IEEE.  Personal use of this material is permitted.  Permission from IEEE must be obtained for all other uses, in any current or future media, including reprinting/republishing this material for advertising or promotional purposes, creating new collective works, for resale or redistribution to servers or lists, or reuse of any copyrighted component of this work in other works.
}
}

\author{\IEEEauthorblockN{Amirreza Mahbod}
\IEEEauthorblockA{Institute for Pathophysiology and \\Allergy Research \\
\textit{Medical University of Vienna}\\
 Vienna, Austria \\
}
\and
\IEEEauthorblockN{Gerald Schaefer}
\IEEEauthorblockA{Department of Computer Science \\
\textit{Loughborough University}\\
Loughborough, United Kingdom}
\and
\IEEEauthorblockN{Chunliang Wang}
\IEEEauthorblockA{Department of Biomedical Engineering \\and Health Systems \\
\textit{KTH Royal Institute of Technology}\\
Stockholm, Sweden}
\and
\IEEEauthorblockN{Rupert Ecker}
\IEEEauthorblockA{\textit{Research and Development Department } \\
\textit{TissueGnostics GmbH}\\
Vienna, Austria \\}
\and
\IEEEauthorblockN{Georg Dorffner}
\IEEEauthorblockA{Section for Artificial Intelligence and \\Decision Support \\
\textit{Medical University of Vienna}\\
Vienna, Austria \\}
\and
\IEEEauthorblockN{Isabella Ellinger}
\IEEEauthorblockA{Institute for Pathophysiology and \\Allergy Research \\
\textit{Medical University of Vienna}\\
 Vienna, Austria \\}
}

\maketitle

\begin{abstract}
Skin cancer is among the most common cancer types. Dermoscopic image analysis improves the diagnostic accuracy for detection of malignant melanoma and other pigmented skin lesions when compared to unaided visual inspection. Hence, computer-based methods to support medical experts in the diagnostic procedure are of great interest. Fine-tuning pre-trained convolutional neural networks (CNNs) has been shown to work well for skin lesion classification. Pre-trained CNNs are usually trained with natural images of a fixed image size which is typically significantly smaller than captured skin lesion images and consequently dermoscopic images are downsampled for fine-tuning. However, useful medical information may be lost during this transformation.

In this paper, we explore the effect of input image size on skin lesion classification performance of fine-tuned CNNs. For this, we resize dermoscopic images to different resolutions, ranging from $64 \times 64$ to $768 \times 768$ pixels and investigate the resulting classification performance of three well-established CNNs, namely DenseNet-121, ResNet-18, and ResNet-50. Our results show that using very small images (of size $64 \times 64$ pixels) degrades the classification performance, while images of size $128 \times 128$ pixels and above support good performance with larger image sizes leading to slightly improved classification.

We further propose a novel fusion approach based on a three-level ensemble strategy that exploits multiple fine-tuned networks trained with dermoscopic images at various sizes. When applied on the ISIC 2017 skin lesion classification challenge, our fusion approach yields an area under the receiver operating characteristic curve of 89.2\% and 96.6\% for melanoma classification and seborrheic keratosis classification, respectively, outperforming state-of-the-art algorithms.
\end{abstract}

\begin{IEEEkeywords}
Dermatology, skin cancer, dermoscopy, medical image analysis, deep learning, image resolution, transfer learning
\end{IEEEkeywords}

\section{Introduction}
Malignant melanoma (MM) is the deadliest type of skin cancer and its incidence rate has increased over the last years~\cite{leiter2014epidemiology}. However, early detection of MM significantly increases the overall survival rate~\cite{schadendorf2018melanoma}. Due to the particular morphological patterns of MM which can resemble other pigmented skin lesions, only 65-80\% of melanomas are correctly diagnosed using unaided visual inspection by experienced medical experts~\cite{brinker2018skin}. 
However, using supportive imaging modalities such as dermoscopy can improve diagnostic accuracy by up to 50\%~\cite{kittler2004dermatoscopy}. In general, diagnostic accuracy highly correlates with expert's experience. Therefore, using a computer-based technique as a tool to support inexperienced physicians or as a second opinion is of great interest. 

The most promising solutions for computer-based skin lesion classification make use of deep learning and in particular of convolutional neural networks (CNNs)~\cite{brinker2018skin}. As the number of publicly available skin lesion images is rather small, transfer learning is the conventional approach to use CNNs for skin lesion classification. Here, pre-trained CNNs are usually fine-tuned to perform the classification task~\cite{mahbod2019fusing, gessert2018skin, brinker2018skin, mahbod2019skin}. Various well-known pre-trained CNNs, such as ResNet~\cite{He2016}, GoogLeNet~\cite{Szegedy2015}, and DenseNet~\cite{huang2017densely}, have been introduced and can be used for skin lesion classification.

Resizing skin lesion images is generally necessary to allow for fine-tuning of pre-trained CNNs since pre-trained CNNs are typically trained with natural images of a fixed image size that is significantly smaller than dermoscopic images. Furthermore, due to computational limitations, it might be impossible to fine-tune networks with original high-resolution skin lesion images. On the other hand, downsampling may lead to a loss of useful medical information and the ideal resizing factor to fine-tune pre-trained CNNs remains an open question. In some previous studies~\cite{Kawahara2016, Yu2017, DeVries2017}, resized skin lesion images larger than the original input size of the utilised CNN were used. However, these studies were limited to a fixed re-scale factor or to a certain CNN. Therefore, the effect of input image size on the skin lesion classification performance still needs to be further explored. 

In this paper, we investigate the effect of image re-scaling on skin lesion classification performance of several fine-tuned CNNs, namely ResNet-18, ResNet-50, and DenseNet-121. We examine the classification performance with re-scaled input images of five different resolutions: $64 \times 64$, $128 \times 128$, $224 \times 224$, $448 \times 448$, and $768 \times 768$ pixels. To our knowledge, this is the first work investigating the effect of using both very small images and very large images on skin lesion classification performance. Moreover, we propose a three-level fusion approach by ensembling the results of different fine-tuned networks that were trained with images at different sizes. Experimental results show this approach to yield the state-of-the-art skin lesion classification performance when applied on the ISIC 2017 challenge test dataset with an average AUC of 92.9\%.

\section{Materials and Methods}
\subsection{Datasets}
We employ two datasets from the ISIC archive\footnote{\url{https://www.isic-archive.com/\#!/topWithHeader/onlyHeaderTop/gallery}}. For fine-tuning pre-trained CNNs, we use the training, validation and test sets of the ISIC 2016 challenge dataset~\cite{gutman2016skin} as well as the training and validation set of the ISIC 2017 challenge dataset~\cite{Codella2017}, which includes three types of skin lesions. From these two datasets, 2187 dermoscopic images are extracted for training, comprising 411 MM, 254 seborrheic keratosis (SK), and 1372 benign nevi (BN) images. For testing our algorithm, we use the 600 test images from the ISIC 2017 challenge dataset. Both training and test images contain various artefacts 
and are of image resolutions ranging from $1022 \times 767$ to $6748 \times 4499$ pixels. 

\subsection{Pre-processing}
\label{preprocess}
For pre-processing, first, we apply a grayworld colour constancy algorithm to normalise the colours of the images as suggested in~\cite{Barata2015}. Then, we subtract the mean intensity RGB value of the ImageNet dataset~\cite{Russakovsky2015} from each individual channel of all training and test images. Finally, we resize the images to the five different resolutions of $64 \times 64$, $128 \times 128$, $224 \times 224$, $448 \times 448$, and $768 \times 768$ pixels using bi-cubic interpolation. For non-square images, the aspect ratio is changed during downsampling.

\subsection{Pre-trained CNNs}
We use three well-established pre-trained CNNs with different depths and architectures, namely ResNet-18~\cite{He2016}, ResNet-50~\cite{He2016} and DenseNet-121~\cite{huang2017densely}. These networks have been shown to give excellent classification performance for various medical image classification tasks including skin lesion classification~\cite{mahbod2018breast, Yu2017, gessert2018skin}. ResNet has a special building block called residual block with skip connections between the input and the output layer in each block. DenseNet's architecture consists of dense blocks that connect each layer to all other layers in a feed-forward fashion. Both ResNet and DenseNet architectures can alleviate the vanishing gradient problem and strengthen feature propagation through the network. Although networks of varying depths exist for both networks, we choose the shallower depths of ResNet-18 and ResNet-50 for the ResNet model and DenseNet-121 for the DenseNet model to prevent overfitting to our limited training data. 

\subsection{Fine-tuning}
\label{fine-tuning}
For fine-tuning, the original fully connected (FC) layers of the pre-trained networks are replaced by two new FC layers with 64 and 3 nodes to adapt to the ternary (MM, SK, BN) classification task similar to~\cite{mahbod2019fusing}. We randomly initialise the weights of these layers from a Gaussian distribution with zero mean and standard deviation of 1. We freeze the initial weight layers of the networks to speed up training and also to prevent overfitting. For DenseNet-121, the dense blocks up to the third block are frozen, while for ResNet-18 and ResNet-50 the residual blocks up to the fourth and the 17th block, respectively, are frozen. All three networks are initially pre-trained on natural images of size $224 \times 224$ pixels. For the other resolutions, we adapt the average pooling layer just before the FC layers to avoid dimensionality mismatch. We examine the effect of training the networks with three different optimisers, namely stochastic gradient descent with momentum (SGDM)~\cite{Murphy2012}, root mean square propagation (RMSProp)~\cite{tieleman2012lecture} and adaptive moment estimation (Adam)~\cite{Kingma2014}. We set the learning rate and momentum to 0.001 and 0.9 for SGDM and the learning rate to 0.0001 for RMSProp and Adam. However, we keep the learning rate of the new FC layers 10 times larger compared to all other weight layers for all networks. We choose varying batch sizes based on the used network, image resolution and the used GPU memory ranging from 16 to 64. Finally, we train the networks for 15 epochs while the learning rate is dropped by a factor of 10 at the fifth and tenth epoch. To artificially increase the training size, we augment training data by image rotations (90, 180 and 270 degrees) and horizontal image flipping, leading to an 8-fold increase of training data. The same augmentation scheme is applied in the inference phase. Thus, for a single test image, rotated and horizontally flipped versions of the test image are fed to the fine-tuned networks and the average result over all 8 augmented images is used for a single test image.

\subsection{Three-level network fusion}
\label{ensemble}
\begin{figure}[t!]
	\centering
	\includegraphics[width=8cm,height=9cm]{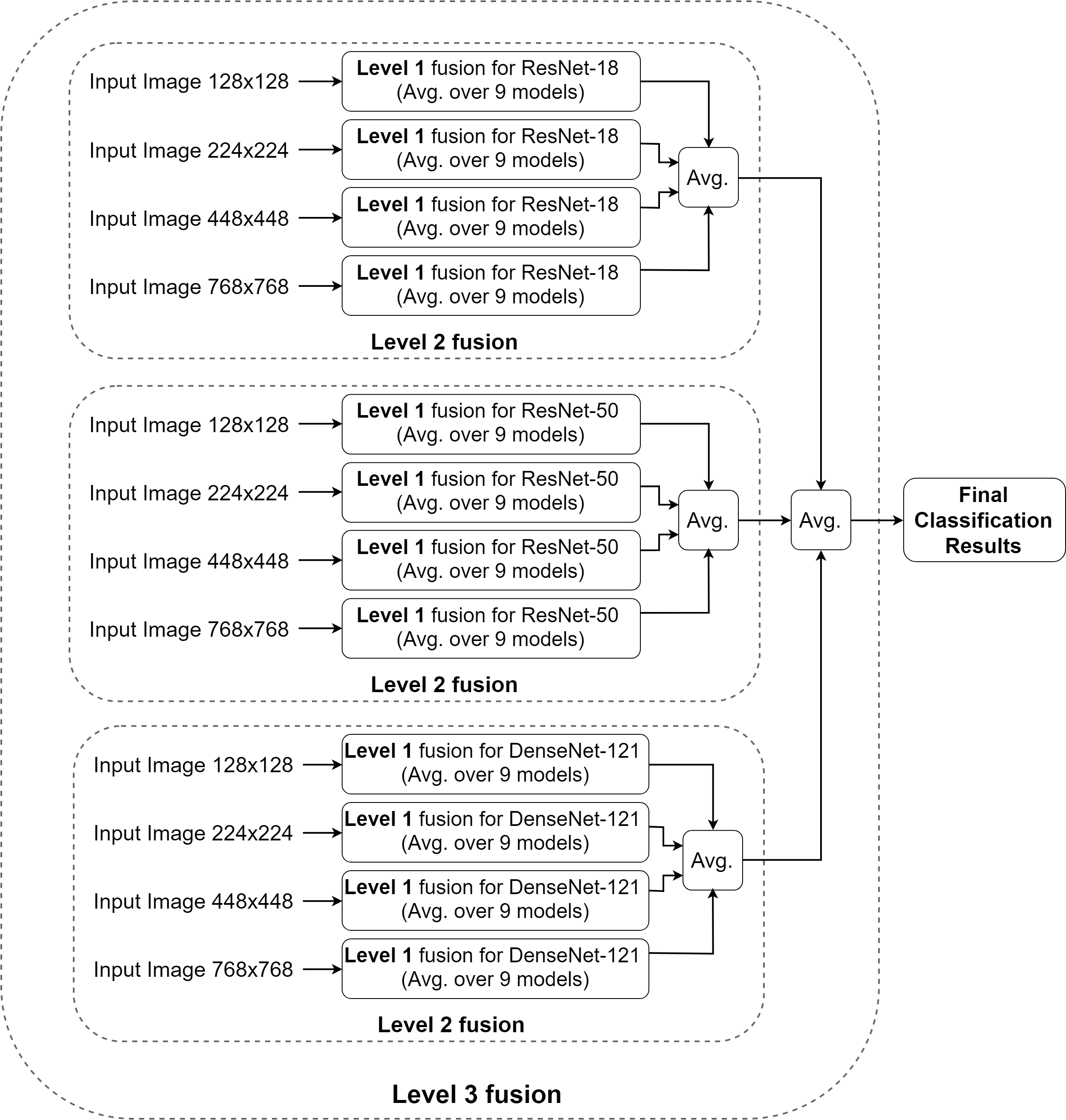} 
	\caption{The proposed three-level fusion approach.}
	\label{3level}
\end{figure}

We develop a three-level fusion scheme which is illustrated in Fig.~\ref{3level}. At level 1, inspired by our earlier work in~\cite{mahbod2019fusing}, we train each network three times with the same hyper-parameters and one optimiser and repeat the procedure for each of the three optimisers. We then take the average over all derived classification probability vectors (i.e., the average over 9 classification outputs for a single network). At level 2, we further fuse the results from the individual networks trained using four different image resolutions (i.e., $128 \times 128$, $224 \times 224$, $448 \times 448$, and $768 \times 768$ pixels). At the third and final fusion level, we fuse the predicted probability vectors of the various architectures to yield the final classification result. The final classification is thus derived from 108 sub-models as shown in Fig.~\ref{3level}.

\begin{table}[b!]
	\centering
	\caption{Effect of inter-architecture network fusion (level 1 fusion) for different models trained on $128 \times 128$ pixel images. Results are given in terms of AUC of the ROC curve [\%], and the results for each optimiser are the averages over three runs.}
	\label{fuse_18}
		\begin{tabular}{llccc}
			\hline \hline
			\textbf{optimiser}  &\textbf{network}& \textbf{MM} & \textbf{SK} & \textbf{avg.} \\
			\hline \hline
			SGDM mean    &ResNet-18 &85.78   &92.99   &89.38  \\
			RMSProp mean &ResNet-18 &84.84   &93.14   &88.99                         \\
			Adam mean    &ResNet-18 &85.15   &92.86   &89.01                      \\
			\hline
			\bf average over optimisers &\bf ResNet-18 &\bf85.46  &\bf93.39    & \bf 89.42  \\
			\hline \hline
			SGDM mean    &ResNet-50 &85.64   & 92.08  & 88.86 \\ 
			RMSProp mean &ResNet-50 &84.74   & 91.50   & 88.12 \\
			Adam mean    &ResNet-50 &83.94   & 92.00  & 87.97 \\
			\hline
			\bf average over optimisers &\bf ResNet-50 &\bf 85.64   &\bf 92.48   & \bf 89.06   \\
			\hline \hline
			SGDM mean    &DenseNet-121 &84.47  &91.64    & 88.06 \\ 
			RMSProp mean &DenseNet-121 &84.84  &93.46    &89.15  \\
			Adam mean    &DenseNet-121 &85.80  &92.92    &89.36 \\
			\hline
			\bf average over optimisers &\bf DenseNet-121 &\bf 85.53  &\bf 93.36   & \bf 89.45  \\
			\hline \hline
		\end{tabular}
\end{table}

\subsection{Evaluation}
As suggested for the ISIC 2017 skin lesion classification challenge, we use the area under the receiver operating characteristics (ROC) curve (AUC) as the main evaluation index. We train all models to solve a ternary classification problem as there are three skin lesion types in the dataset. However, as the ISIC 2017 challenge evaluation is based on two binary classification tasks, namely MM vs.\ all and SK vs.\ all, we convert the three elementary prediction vectors to two elementary prediction vectors using a one-versus-all approach. This allows us to compare our results with other algorithms previously applied on the same dataset. 

\section{Results}
The trained models are evaluated on the (unseen) 600 test images of the ISIC 2017 challenge for skin lesion classification. In particular, there are 117 MMs, 90 SKs, and 393 BN dermoscopic images. We use identical pre-processing and augmentation techniques for all test images as described in Section~\ref{preprocess} and Section~\ref{fine-tuning}.

As described in Section~\ref{ensemble}, to obtain a more robust and improved classification performance for each individual network and for each image resolution, we fuse the results of 9 models (level 1 fusion in Fig.~\ref{3level}). The results obtained by this fusion scheme for (as an example) images of $128 \times 128$ pixels are given in Table~\ref{fuse_18}.

Next, we investigate the effect of input image sizes on the classification performances of the various fine-tuned deep models. The results of this are given in Table~\ref{size} for the five image sizes with the results of each network obtained by level 1 fusion from 9 models as explained above.

\begin{table}[t!]
	\centering
	\caption{Effect of input image size on the classification performance of fine-tuned networks (based on level 1 fusion). Results are given in terms of AUC of the ROC curve [\%].}
	\label{size}
		\begin{tabular}{lcccc}
			\hline \hline
			\textbf{network}  &\textbf{input size}& \textbf{MM} & \textbf{SK} & \textbf{avg.} \\
			\hline \hline
			ResNet-18               & $64\times64$  &78.86  &89.55  & 84.21  \\
			ResNet-50               & $64\times64$  &78.44  &87.54  &82.99   \\
			DenseNet-121            & $64\times64$  &79.21  &88.29  &83.75  \\
			\hline
			ResNet-18               & $128\times128$  &85.46 &93.39 &89.42  \\
			ResNet-50               & $128\times128$  &85.64 &92.48 &89.06  \\
			DenseNet-121            & $128\times128$  &85.53 &93.36 &89.45   \\
			\hline
			ResNet-18               & $224\times224$ &85.37 &93.81 &89.59   \\
			ResNet-50               & $224\times224$ &85.06 &92.77 &88.92 \\
			DenseNet-121            & $224\times224$ &86.30 &93.13 &89.72   \\
			\hline
			ResNet-18               & $448\times448$ &89.20 &95.54 &92.37 \\
			ResNet-50               & $448\times448$ &85.58 &95.03 &90.31\\
			DenseNet-121            & $448\times448$ &86.11 &93.41 &89.76 \\
			\hline
			ResNet-18               & $768\times768$ &88.89 &95.85 &92.37   \\
			ResNet-50               & $768\times768$ &88.70 &95.64 &92.17 \\	
			DenseNet-121            & $768\times768$ &85.43 &94.16 &89.80  \\
			\hline \hline
		\end{tabular}
\end{table}

Table~\ref{fusion} shows the results obtained by the higher-level fusion schemes (i.e., level 2 and level 3 fusion in Fig.~\ref{3level}) as described in Section~\ref{ensemble}. We exclude he smallest image size ($64\times64$ pixels) since, as is apparent from Table~\ref{size}, these led to significantly degraded classification performance.

\begin{table}[b!]
	\caption{Effect of level two and level three fusion schemes on the classification performance. Results are given in terms of AUC of the ROC curve [\%]}
	\centering
	\label{fusion}
		\begin{tabular}{lcccc}
			\hline \hline
			\textbf{network}  &\textbf{input size}& \textbf{MM} & \textbf{SK} & \textbf{avg.} \\
			\hline \hline
			ResNet-18 (level 2)     &  all sizes   &89.12 &96.26 &92.69\\
			ResNet-50 (level 2)    &  all sizes   &88.50 &96.03 &92.27\\
			 DenseNet-121 (level 2)  &  all sizes   &87.69 &95.77 &91.73 \\
			\hline
			level 3 fusion        &  all sizes   &89.16 &96.57 &92.86\\
			\hline \hline
		\end{tabular}
\end{table}

We also evaluate another fusion strategy which performs fusion of the three networks' outputs for a single image resolution and compare this with the proposed three-level fusion approach. The results of this comparison are shown in Table~\ref{single fusion}.

\begin{table}[t!]
	\caption{Results of fusing different networks for a singe image resolution and comparison with the proposed three-level fusion scheme. Results are given in terms of AUC of the ROC curve [\%]}
	\centering
	\label{single fusion}
		\begin{tabular}{lcccc}
			\hline \hline
			\textbf{network}  &\textbf{input size}& \textbf{MM} & \textbf{SK} & \textbf{avg.} \\
			\hline \hline
			fusion of all nets     & $64\times64$     &80.59 &89.67 &85.13\\
			fusion of all nets     & $128\times128$   &86.39 &94.26 &90.32\\
			fusion of all nets     & $224\times224$   &86.86 &94.42 &90.64\\
			fusion of all nets     & $448\times448$   &88.99 &96.05 &92.52 \\
			fusion of all nets     & $768\times768$   &88.70 &95.64 &92.17 \\
			\hline
			three-level fusion          &  all sizes   &89.16 &96.57 &92.86\\
			\hline \hline
		\end{tabular}
\end{table}

\begin{figure}[b!]
	\centering
	\includegraphics[width=\columnwidth]{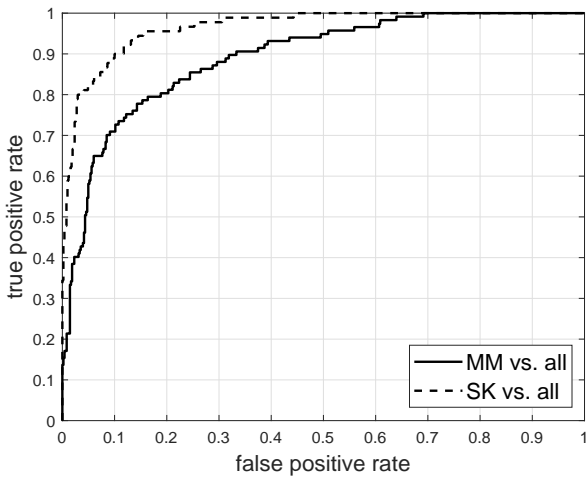}
	\caption{ROC curve for the proposed three-level fusion approach.}
	\label{ROC}
\end{figure}

Fig.~\ref{ROC} shows the receiver operating characteristic curve (ROC) of the best approach which combines the results from all fusion levels. 

\begin{figure}[t!]
	\centering
	\includegraphics[width=\columnwidth]{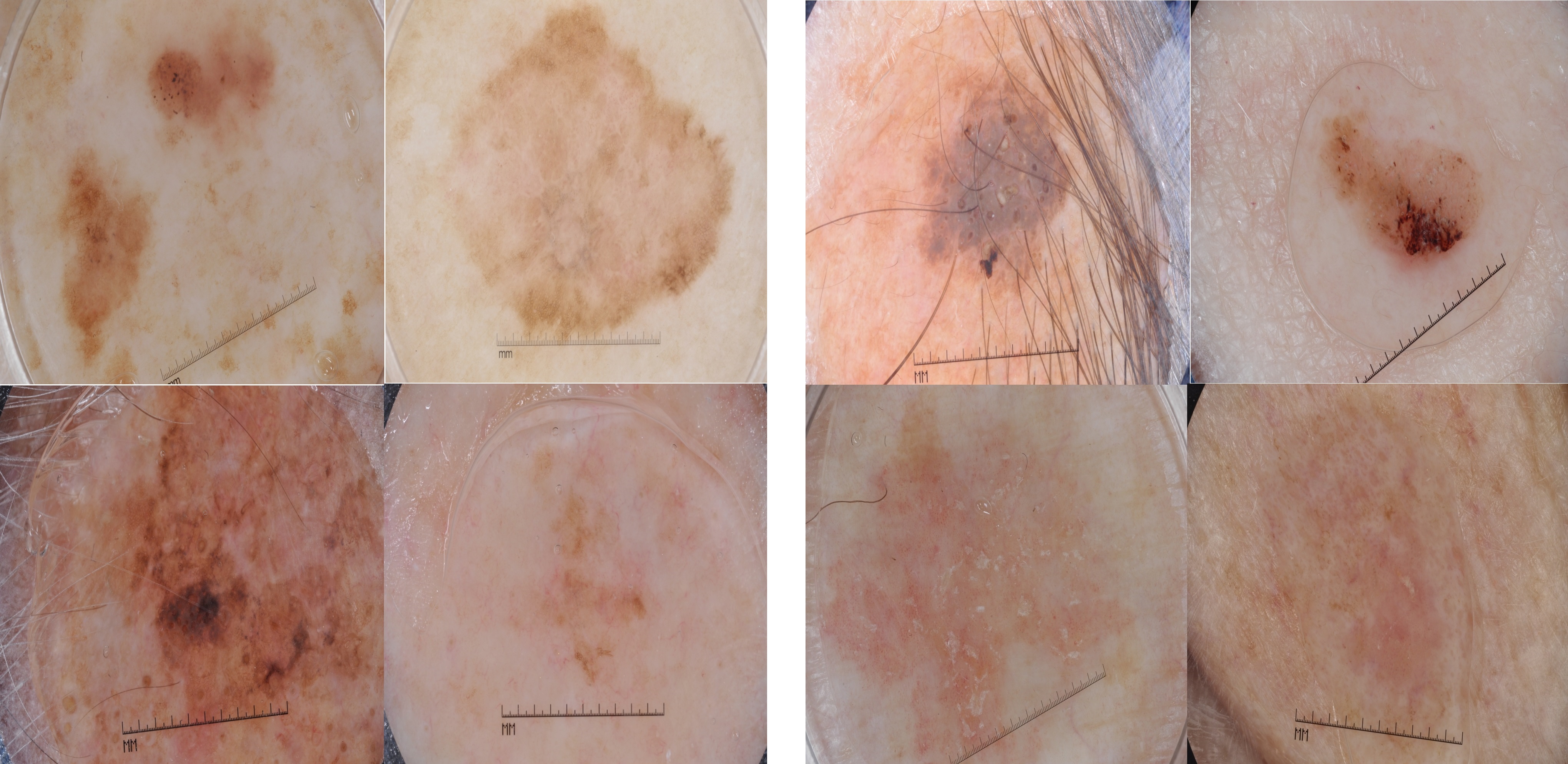}
	\caption{Examples of correctly classified images for MM vs.\ all (left) and SK vs.\ all (right) tasks.}
	\label{correct}
\end{figure}

\begin{figure}[t!]
	\centering
	\includegraphics[width=\columnwidth]{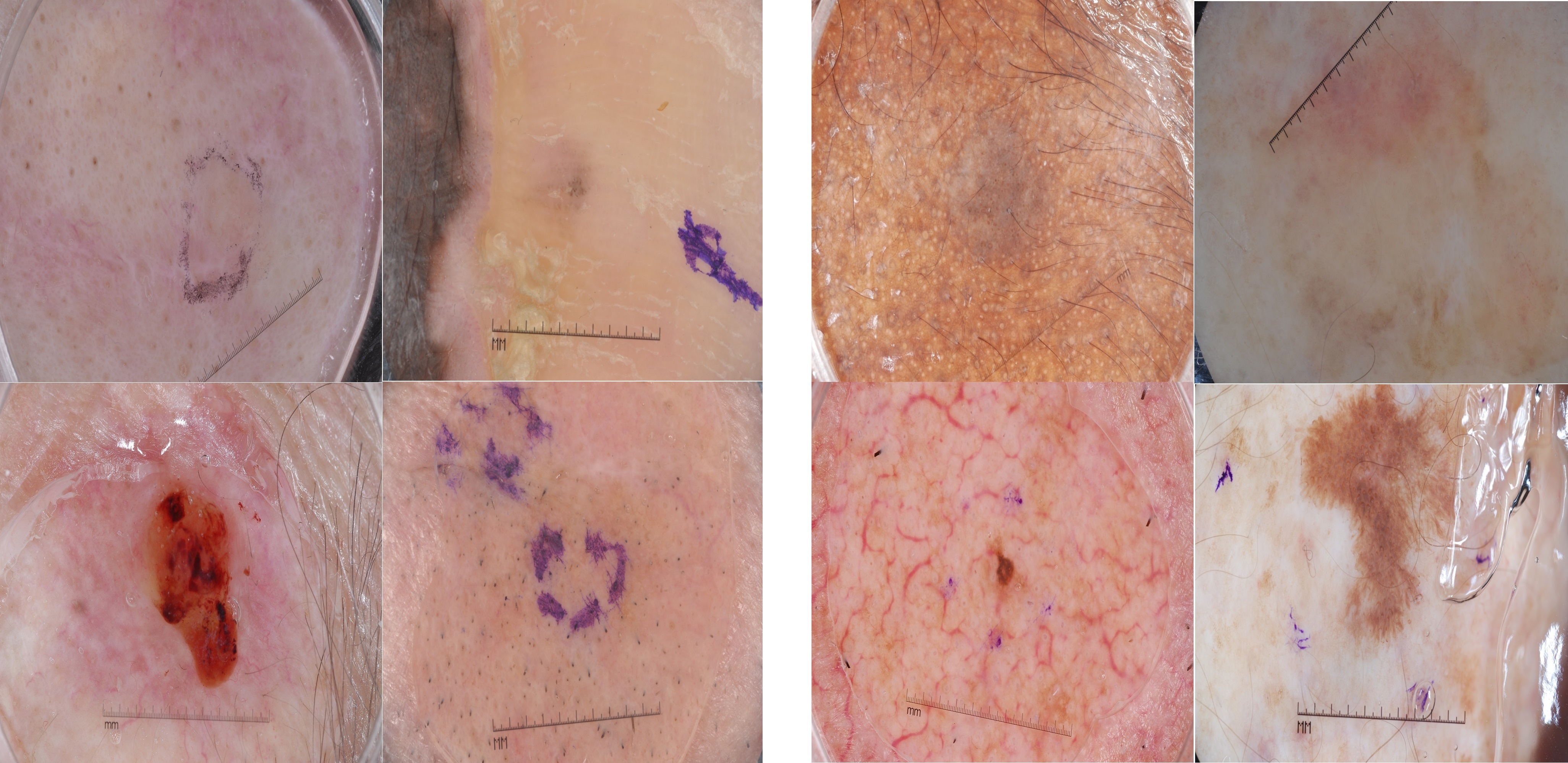}
	\caption{Examples of incorrectly classified images for MM vs.\ all (left) and SK vs.\ all (right) tasks.}
	\label{incorrect}
\end{figure}

Fig.~\ref{correct} and Fig.~\ref{incorrect} give examples of correctly and incorrectly classified skin lesion images of our three-level fusion approach, respectively. For these, to convert the three elementary prediction vectors, we choose the highest probability as the predicted class by the model in each prediction vector.

We compare the performance of our proposed fusion approach (i.e., last row in Table~\ref{fusion}) with the top three performers of the ISIC 2017 competition as well as with other state-of-the-art algorithms which have been applied on the same dataset and have shown superior classification performance compared to the ISIC 2017 challenge winners. The results of the comparison are given in Table~\ref{tab:comparison} in terms of AUC.

\begin{table}[b!]
\caption{Comparison of the ISIC 2017 challenge winners (rows 1-3), additional state-of-the-art methods (rows 4-7), and our proposed approach (final row) based on AUC scores [\%].}
\centering
\label{tab:comparison}
\begin{tabular}{lcccc}
\hline \hline
\textbf{approach}  &\textbf{input size}& \textbf{MM} & \textbf{SK} & \textbf{avg.} \\
\hline \hline
Matsunaga {\it et al.}~\cite{Matsunaga2017}  & n/a &86.8 &95.3 &91.1\\
Gonzalez-Diaz~\cite{Diaz2017}                & $256\times256$  &85.6 &96.5 &91.0\\
Menegola {\it et al.}~\cite{Menegola2017}    & $128\times128$  &87.4 &94.3 &90.8\\
\hline
Mahbod {\it et al.}~\cite{mahbod2019fusing}  & $224\times224$ &87.3 &95.5 &91.4\\
Zhang {\it et al.}~\cite{zhang2019attention} & $224\times224$ &87.5 &95.8 &91.7\\
Yan {\it et al.}~\cite{10.1007/978-3-030-20351-1_62} & $256\times256$  &88.3 & n/a & n/a\\
Guo {\it et al.}~\cite{GUO201867} & $224\times224$  &87.4 & 95.9 & 91.7\\
\hline
three-level fusion & multiple & 89.2 & 96.6 &92.9\\
\hline \hline
\end{tabular}
\end{table}

Matsunaga {\it et al.}~\cite{Matsunaga2017}, top-ranked in the competition, made use of colour constancy as a pre-processing step and two separate classifiers for the two binary classification problems (i.e., MM vs.\ all and SK vs.\ all). For each classifier, a fine-tuned ResNet-50 was used as backbone model. As post-processing step, sex and age information were fused with the classifier's output to yield final classification results. The down-sampling factor was not reported in their approach. Gonzalez-Diaz~\cite{Diaz2017}, the runner up, performed classification in a multi-step approach that employed three deep models including a full CNN to segment lesion areas in the images, a constrained CNN to add more clinical features for better categorisation, and a modified fine-tuned ResNet-50 to perform final classification. They used images resized to $256\times256$ pixels for network training. Menegola {\it et al.}~\cite{Menegola2017}, the third-ranked team, used extensive external data sources from the ISIC archive and ensembled seven fine-tuned models (six models based on the Inception-v4 architecture~\cite{Szegedy2017} and one model based on the ResNet-101 architecture). Images down-sampled to $128\times128$ pixels were used which made the algorithm faster compared to the other winners of the competition.

Several methods were developed after the ISIC 2017 competition and reported better performance compared to the challenge winners. In our earlier work~\cite{mahbod2019fusing}, we used inter and intra-architecture network fusion to extract deep features from several fine-tuned deep models. However, a single image resolution of $224\times224$ pixels was used in this approach. Zhang {\it et al.}~\cite{zhang2019attention} proposed a novel attention residual learning CNN whose residual blocks aim to prevent the degradation problem and with an attention mechanism to force the network to focus on lesion areas. A pre-trained ResNet-50 model served as backbone model and image patches of size $224\times224$ pixels acquired by central cropping of the original images at different scales were used. Similar to~\cite{zhang2019attention}, Yan {\it et al.}~\cite{10.1007/978-3-030-20351-1_62} also used the idea of a learnable attention mechanism. They added two attention modules to a pre-trained VGG16 network and concatenated the features from the attention modules and the last convolutional layer by an average pooling layer before performing classification. Images down-sampled to $224\times224$ pixels were used in their study. Guo {\it et al.}~\cite{GUO201867} proposed a multi-channel ResNet to perform skin lesion classification. In their approach, an OverFeat model~\cite{sermanet2013overfeat} was used to crop skin lesion images. Then, the images were pre-processed by various techniques and used to fine-tune a number of ResNet-50 models. The image features from different models were concatenated and were used to perform skin lesion classification. Similar to other aforementioned methods, all images were resized to a fixed image size of $224\times224$ pixels in their method. 


Our algorithm is implemented in MatLab (ver. 2018a) based on the MatConvNet framework and the MatLab Neural Network Toolbox. All experiments were conducted on a single workstation with an Intel Corei5-6600k 3.50 GHz CPU, 16 GB of RAM and a single nVIDIA GTX 1070 card with 8 GB of installed memory. The average training times for each deep architecture and each image resolution are reported in Table~\ref{time}. The training times for the different optimisers vary slightly and the reported results in Table~\ref{time} are the average training times in minutes. 

\begin{table}[t!]
	\centering
	\caption{Average training times (in minutes) for each individual model and image resolution.}
	\label{time}
	\begin{tabular}{cccc}
		\hline
		image size & ResNet-18 & ResNet-50 &  DenseNet-121 \\
		\hline
		$64\times64$   & 102  & 140  & 302  \\
		$128\times128$ & 110  & 177  & 355  \\
		$224\times224$ & 155  & 230 &760  \\
		$448\times448$ & 197  & 320 & 880 \\
		$768\times768$ & 520  & 825 & 2150  \\
		\hline
	\end{tabular}	
\end{table}

\section{Discussion}
In this study, we explicitly investigate the effect of image re-scaling on skin lesion classification performance of several CNNs. Moreover, we achieve state-of-the-art classification performance on the ISIC 2017 challenge dataset by proposing a straightforward three-level ensemble strategy that uses multiple fine-tuned CNNs and multi-resolution dermoscopic images. 


From the results in Table~\ref{fuse_18}, we can see that fine-tuning pre-trained network models with different optimisers deliver comparable classification performance. However, combining the results from the different optimisers leads to a better average AUC compared to the individual AUCs of all three models. This fusion step was inspired by our earlier work in~\cite{mahbod2019fusing}, where a combination of 18 models was used for inter-network fusion. In contrast to there, here, we just fuse the results of nine models to reduce training time (the other nine models in~\cite{mahbod2019fusing} were trained by the same parameters, but with a different pre-processing step).

The results in Table~\ref{size} show the effect of input image size on the classification performance. The obtained results are of interest for several reasons. First, down-sampled images, even at a drastically reduced image size of $64\times64$ pixels, still hold valuable information for classification as even the lowest AUC obtained (82.99\%) is useful. However, the general performance of the models trained on  $64\times64$ pixel images is significantly lower compared to the results obtained using images with higher resolution. Thus, it is evident that heavy downsampling causes a loss of valuable information, which is also the reason we exclude the lowest image resolution from the subsequent fusion schemes.

Second, Table~\ref{size} shows a tendency of improved classification performance with increasing image size. The average results over the  different models in Table~\ref{size} are 89.31\%, 89.41\%, 90.81\% and 91.44\% for input image resolutions of $128\times128$, $224\times224$, $448\times448$ and $768\times768$ pixels, respectively. If we fuse the results from all three networks of a specific image resolution (i.e., level 2 fusion), we obtain improved average AUC values of 90.32\%, 90.64\%, 92.52\% and 92.52\%, respectively. Thus, in both cases, an increase in image resolution also leads to an improvement in classification accuracy. Since the smallest image resolution in the dataset was $1022 \times768$ pixels and also considering computational limitations, we did not conduct further experiments with images larger than $768 \times768$. To our knowledge, fine-tuning any model for skin lesion classification with resized images of $768 \times768$ pixels is performed for the first time in this work.

Third, Table~\ref{size} allows for a comparison of the individual performances of the employed fine-tuned network models. Considering the average performance of each model for various image resolutions (i.e., 90.93\%, 90.11\% and 89.68\% for ResNet-18, ResNet-50 and DenseNet-121, respectively), ResNet-18 shows the best performance. The classification performance of the same models in the ImageNet Large Scale Visual Recognition Challenge~\cite{Russakovsky2015} is reversed (i.e., DenseNet121 is the best and ResNet-18 the worst). However, considering the number of training examples of the utilised dataset, we can infer that deeper models such as DenseNet-121 have a greater potential to overfit to the small training data size in this study, while shallower models such as ResNet-18 generalise better.

The results in Table~\ref{fusion} show the effect of the second- and third-level fusion schemes of our approach. From there, it is apparent that the multi-resolution fusion approach delivers a better classification performance compared to any single image resolution network. Our proposed three-level fusion approach, which combines the results of 108 models is shown to yield better classification performance still, outperforming all single networks and all lower level fusion schemes.

The results in Table~\ref{single fusion} show that fusing different networks at a single image resolution can also lead to improved classification. In general, the fusion of the three network models, i.e.\ ResNet-18, ResNet-50 and DenseNet-121 gives better results compared to any single network although, interestingly, this is not the case for the highest image resolution of $768 \times 768$ pixels where ResNet-18 performs slightly better than the combination of the three networks. Our proposed three-level approach is superior compared to all networks fused this way.

The comparative results in Table~\ref{tab:comparison} show that our proposed fusion scheme outperforms other state-of-the-art algorithms for both MM and SK recognition and with an improvement of at least 1.2\% in terms of average AUC, confirming it to be a powerful approach for skin lesion classification. 

While all reported results are derived from the ISIC 2017 challenge test dataset, a direct comparison of the classification performance is not trivial as different training sets were used in the different approaches. However, our method exploits fewer external training samples compared to most of the other approaches; 1444 external training samples were used in~ \cite{Matsunaga2017}, 900 in~\cite{Diaz2017}, 7544 in~\cite{Menegola2017}, and 1320 in~\cite{zhang2019attention}, while we utilised only 187 external training images (the same number as in~\cite{mahbod2019fusing}).

Looking at Table~\ref{time}, it is apparent that the training time required for DenseNet-121 is higher compared to the other networks, which was expected since this architecture is significantly deeper than the other ones. Also as expected, more time is required for training networks with higher resolution images since more convolutions need to be performed in each layer of the networks. 


Finally, Fig.~\ref{correct} illustrates that our proposed algorithm is able to correctly classify challenging skin lesion images that contain various artefacts such as skin hair or ruler charts as well as images that would be difficult to automatically segment correctly. From Fig.~\ref{incorrect} we can see that some even more challenging images are still misclassified, including some images where the lesion borders are not well defined and samples where the lesion occupies only a small part of the image.

While we have evaluated the effects of input image sizes as well as the effects of multi-model and multi-resolution image fusion for skin lesion classification, there are some limitations in this work. The biggest limitation of our proposed fusion approach is the training time required to derive the classification models which may not be suitable for application in a clinical setting. However, as it is possible to train different networks in parallel, the overall training time can be significantly reduced by accessing a number of suitable computational devices. Another consideration of our ensembling method is the fusion scheme which is averaging. In~\cite{godinez2017multi}, a multi-scale CNN (M-CNN) was proposed that used multiple scale images in one single network. However, as the network width increased drastically, they could only use three convolutional layers which led to a very shallow network. Moreover, with a new architecture proposed, they had to train the model from scratch and hence were unable to take advantage of transfer learning, while their approach also did not allow to evaluate the contribution of each image scale to the final classification performance. However, with sufficient computational power, the classification performance of an M-CNN with pre-trained deep models for each image scale can be investigated. Another issue that can be further addressed in future work is the resizing factor. While in this paper we utilised five downsampling factors, the effect of other image resolutions between the minimum and maximum sizes can also be investigated. Finally, the number of pre-trained networks that we use is limited to three pre-trained CNNs. Exploring other architectures may be addressed in future studies.

\section{Conclusions} 
In this paper, we have investigated the effect of image resolutions for transfer learning classification performance in the context of skin lesion analysis. The results of our study show that while down-sampling images to a very low resolution may not be optimal for fine-tuning pre-trained convolutional neural networks, even low-resolution images yield acceptable classification results. In contrast, images with higher resolution support further improved classification performance. In addition, we have presented a three-level fusion approach that combines results from different networks and different image resolutions and is demonstrated to result in the best classification performance compared to a number of state-of-the-art algorithms for skin lesion analysis and evaluated on the ISIC 2017 skin lesion classification challenge dataset. 

\section*{Acknowledgements}
This research has received funding from the Marie Sklodowska-Curie Actions of the European Union's Horizon 2020 programme under REA grant agreement no. 675228. The authors thank nVIDIA corporation for their generous GPU donation.

\balance
\bibliographystyle{IEEEtran}
\bibliography{icpr20}

\begin{thebibliography}{10}
\providecommand{\url}[1]{#1}
\csname url@samestyle\endcsname
\providecommand{\newblock}{\relax}
\providecommand{\bibinfo}[2]{#2}
\providecommand{\BIBentrySTDinterwordspacing}{\spaceskip=0pt\relax}
\providecommand{\BIBentryALTinterwordstretchfactor}{4}
\providecommand{\BIBentryALTinterwordspacing}{\spaceskip=\fontdimen2\font plus
\BIBentryALTinterwordstretchfactor\fontdimen3\font minus
  \fontdimen4\font\relax}
\providecommand{\BIBforeignlanguage}[2]{{%
\expandafter\ifx\csname l@#1\endcsname\relax
\typeout{** WARNING: IEEEtran.bst: No hyphenation pattern has been}%
\typeout{** loaded for the language `#1'. Using the pattern for}%
\typeout{** the default language instead.}%
\else
\language=\csname l@#1\endcsname
\fi
#2}}
\providecommand{\BIBdecl}{\relax}
\BIBdecl

\bibitem{leiter2014epidemiology}
U.~Leiter, T.~Eigentler, and C.~Garbe, ``Epidemiology of skin cancer,'' in
  \emph{Sunlight, Vitamin D and Skin Cancer}.\hskip 1em plus 0.5em minus
  0.4em\relax Springer, 2014, pp. 120--140.

\bibitem{schadendorf2018melanoma}
D.~Schadendorf, A.~C. van Akkooi, C.~Berking, K.~G. Griewank, R.~Gutzmer,
  A.~Hauschild, A.~Stang, A.~Roesch, and S.~Ugurel, ``Melanoma,'' \emph{The
  Lancet}, vol. 392, no. 10151, pp. 971--984, 2018.

\bibitem{brinker2018skin}
T.~J. Brinker, A.~Hekler, J.~S. Utikal, N.~Grabe, D.~Schadendorf, J.~Klode,
  C.~Berking, T.~Steeb, A.~H. Enk, and C.~von Kalle, ``Skin cancer
  classification using convolutional neural networks: Systematic review,''
  \emph{Journal of Medical Internet Research}, vol.~20, no.~10, p. e11936,
  2018.

\bibitem{kittler2004dermatoscopy}
H.~Kittler, ``Dermatoscopy of pigmented skin lesions,'' \emph{Giornale Italiano
  di Dermatologia e Venereologia}, vol. 139, no.~6, pp. 541--546, 2004.

\bibitem{mahbod2019fusing}
A.~Mahbod, G.~Schaefer, I.~Ellinger, R.~Ecker, A.~Pitiot, and C.~Wang, ``Fusing
  fine-tuned deep features for skin lesion classification,'' \emph{Computerized
  Medical Imaging and Graphics}, vol.~71, pp. 19--29, 2019.

\bibitem{gessert2018skin}
N.~Gessert, T.~Sentker, F.~Madesta, R.~Schmitz, H.~Kniep, I.~Baltruschat,
  R.~Werner, and A.~Schlaefer, ``Skin lesion diagnosis using ensembles,
  unscaled multi-crop evaluation and loss weighting,'' \emph{arXiv preprint
  arXiv:1808.01694}, 2018.

\bibitem{mahbod2019skin}
A.~Mahbod, G.~Schaefer, C.~Wang, R.~Ecker, and I.~Ellinger, ``Skin lesion
  classification using hybrid deep neural networks,'' in \emph{International
  Conference on Acoustics, Speech and Signal Processing}.\hskip 1em plus 0.5em
  minus 0.4em\relax IEEE, May 2019, pp. 1229--1233.

\bibitem{He2016}
K.~He, X.~Zhang, S.~Ren, and J.~Sun, ``Deep residual learning for image
  recognition,'' in \emph{Conference on Computer Vision and Pattern
  Recognition}.\hskip 1em plus 0.5em minus 0.4em\relax IEEE, 2016, pp.
  770--778.

\bibitem{Szegedy2015}
C.~Szegedy, W.~Liu, Y.~Jia, P.~Sermanet, S.~Reed, D.~Anguelov, D.~Erhan,
  V.~Vanhoucke, and A.~Rabinovich, ``Going deeper with convolutions,'' in
  \emph{Proceedings of the IEEE Conference on Computer Vision and Pattern
  Recognition}.\hskip 1em plus 0.5em minus 0.4em\relax IEEE, 2015, pp. 1--9.

\bibitem{huang2017densely}
G.~Huang, Z.~Liu, L.~Van Der~Maaten, and K.~Q. Weinberger, ``Densely connected
  convolutional networks.'' in \emph{Proceedings of the IEEE Conference on
  Computer Vision and Pattern Recognition}, vol.~1, no.~2.\hskip 1em plus 0.5em
  minus 0.4em\relax IEEE, 2017, pp. 4700--4708.

\bibitem{Kawahara2016}
J.~Kawahara, A.~BenTaieb, and G.~Hamarneh, ``Deep features to classify skin
  lesions,'' in \emph{International Symposium on Biomedical Imaging}.\hskip 1em
  plus 0.5em minus 0.4em\relax IEEE, 2016, pp. 1397--1400.

\bibitem{Yu2017}
Z.~Yu, X.~Jiang, T.~Wang, and B.~Lei, ``Aggregating deep convolutional features
  for melanoma recognition in dermoscopy images,'' in \emph{International
  Workshop on Machine Learning in Medical Imaging}.\hskip 1em plus 0.5em minus
  0.4em\relax Springer, 2017, pp. 238--246.

\bibitem{DeVries2017}
T.~DeVries and D.~Ramachandram, ``Skin lesion classification using deep
  multi-scale convolutional neural networks,'' \emph{arXiv preprint
  arXiv:1703.01402}, 2017.

\bibitem{gutman2016skin}
D.~Gutman, N.~C. Codella, E.~Celebi, B.~Helba, M.~Marchetti, N.~Mishra, and
  A.~Halpern, ``Skin lesion analysis toward melanoma detection: A challenge at
  the {I}nternational {S}ymposium on {B}iomedical {I}maging ({ISBI}) 2016,
  hosted by the {I}nternational {S}kin {I}maging {C}ollaboration ({ISIC}),''
  \emph{arXiv preprint arXiv:1605.01397}, 2016.

\bibitem{Codella2017}
N.~C.~F. Codella, D.~Gutman, M.~E. Celebi, B.~Helba, M.~A. Marchetti, S.~W.
  Dusza, A.~Kalloo, K.~Liopyris, N.~Mishra, and H.~Kittler, ``Skin lesion
  analysis toward melanoma detection: A challenge at the 2017 {I}nternational
  {S}ymposium on {B}iomedical {I}maging ({ISBI}), hosted by the {I}nternational
  {S}kin {I}maging {C}ollaboration ({ISIC}),'' \emph{arXiv preprint
  arXiv:1710.05006}, 2017.

\bibitem{Barata2015}
C.~Barata, M.~E. Celebi, and J.~S. Marques, ``Improving dermoscopy image
  classification using color constancy,'' \emph{IEEE Journal of Biomedical and
  Health Informatics}, vol.~19, no.~3, pp. 1146--1152, 2015.

\bibitem{Russakovsky2015}
O.~Russakovsky, J.~Deng, H.~Su, J.~Krause, S.~Satheesh, S.~Ma, Z.~Huang,
  A.~Karpathy, A.~Khosla, and M.~Bernstein, ``{ImageNet} large scale visual
  recognition challenge,'' \emph{International Journal of Computer Vision},
  vol. 115, no.~3, pp. 211--252, 2015.

\bibitem{mahbod2018breast}
A.~Mahbod, I.~Ellinger, R.~Ecker, {\"O}.~Smedby, and C.~Wang, ``Breast cancer
  histological image classification using fine-tuned deep network fusion,'' in
  \emph{Image Analysis and Recognition}, A.~Campilho, F.~Karray, and B.~ter
  Haar~Romeny, Eds.\hskip 1em plus 0.5em minus 0.4em\relax Cham: Springer
  International Publishing, 2018, pp. 754--762.

\bibitem{Murphy2012}
K.~P. Murphy, \emph{Machine Learning: A Probabilistic Perspective}.\hskip 1em
  plus 0.5em minus 0.4em\relax MIT press, 2012.

\bibitem{tieleman2012lecture}
T.~Tieleman and G.~Hinton, ``Lecture 6.5-rmsprop: Divide the gradient by a
  running average of its recent magnitude,'' \emph{COURSERA: Neural networks
  for machine learning}, vol.~4, no.~2, pp. 26--31, 2012.

\bibitem{Kingma2014}
D.~P. Kingma and J.~Ba, ``Adam: A method for stochastic optimization,''
  \emph{arXiv preprint arXiv:1412.6980}, 2014.

\bibitem{Matsunaga2017}
K.~Matsunaga, A.~Hamada, A.~Minagawa, and H.~Koga, ``Image classification of
  melanoma, nevus and seborrheic keratosis by deep neural network ensemble,''
  \emph{arXiv preprint arXiv:1703.03108}, 2017.

\bibitem{Diaz2017}
I.~G. D{\'{i}}az, ``Incorporating the knowledge of dermatologists to
  convolutional neural networks for the diagnosis of skin lesions,''
  \emph{arXiv preprint arXiv:1703.01976}, 2017.

\bibitem{Menegola2017}
A.~Menegola, J.~Tavares, M.~Fornaciali, L.~T. Li, S.~Avila, and E.~Valle,
  ``{RECOD} titans at {ISIC} challenge 2017,'' \emph{arXiv preprint
  arXiv:1703.04819}, 2017.

\bibitem{zhang2019attention}
J.~Zhang, Y.~Xie, Y.~Xia, and C.~Shen, ``Attention residual learning for skin
  lesion classification,'' \emph{IEEE Transactions on Medical Imaging},
  vol.~38, no.~9, pp. 2092--2103, 2019.

\bibitem{10.1007/978-3-030-20351-1_62}
Y.~Yan, J.~Kawahara, and G.~Hamarneh, ``Melanoma recognition via visual
  attention,'' in \emph{Information Processing in Medical Imaging}, A.~C.~S.
  Chung, J.~C. Gee, P.~A. Yushkevich, and S.~Bao, Eds.\hskip 1em plus 0.5em
  minus 0.4em\relax Cham: Springer International Publishing, 2019, pp.
  793--804.

\bibitem{GUO201867}
S.~Guo and Z.~Yang, ``Multi-channel-{ResNet}: An integration framework towards
  skin lesion analysis,'' \emph{Informatics in Medicine Unlocked}, vol.~12, pp.
  67 -- 74, 2018.

\bibitem{Szegedy2017}
C.~Szegedy, S.~Ioffe, V.~Vanhoucke, and A.~A. Alemi, ``{I}nception-v4,
  {I}nception-{R}esnet and the impact of residual connections on learning,'' in
  \emph{Association for the Advancement of Artificial Intelligence}, 2017, pp.
  4278--4284.

\bibitem{sermanet2013overfeat}
P.~Sermanet, D.~Eigen, X.~Zhang, M.~Mathieu, R.~Fergus, and Y.~LeCun,
  ``{OverFeat}: Integrated recognition, localization and detection using
  convolutional networks,'' \emph{arXiv preprint arXiv:1312.6229}, 2013.

\bibitem{godinez2017multi}
W.~J. Godinez, I.~Hossain, S.~E. Lazic, J.~W. Davies, and X.~Zhang, ``A
  multi-scale convolutional neural network for phenotyping high-content
  cellular images,'' \emph{Bioinformatics}, vol.~33, no.~13, pp. 2010--2019,
  2017.

\end{thebibliography}

\end{document}